\PassOptionsToPackage{numbers,sort&compress}{natbib}

\documentclass{article}

\usepackage[main, final]{styles/neurips_2026}

\usepackage[utf8]{inputenc} 
\usepackage[T1]{fontenc}    
\usepackage{hyperref}       
\usepackage{url}            
\usepackage{booktabs}       
\usepackage{amsfonts}       
\usepackage{nicefrac}       
\usepackage{microtype}      
\usepackage{xcolor}         

\usepackage{amsmath,amssymb,bm}
\usepackage{array}
\usepackage{float}
\usepackage{multirow}
\usepackage{subcaption}
\usepackage{threeparttable}
\usepackage[dvipsnames]{xcolor}
\usepackage[ruled,noline,noend]{algorithm2e}
\usepackage{siunitx}
\usepackage{tikz}
\usetikzlibrary{shapes, arrows, arrows.meta, positioning, fit, backgrounds, calc, decorations.pathreplacing}
\hypersetup{
  colorlinks = true,
  linkcolor={green!80!black},
  citecolor={red!70!black},
  urlcolor={blue!70!black}
}

\newcommand{\wt}[1]{\tilde{#1}}

\newcommand{\mcal}[1]{\mathcal{#1}}
\newcommand{\mrm}[1]{\mathrm{#1}}
\newcommand{\mbb}[1]{\mathbb{#1}}
\newcommand{\mb}[1]{\mathbf{#1}}

\newcommand{\amul}{\odot}

\DontPrintSemicolon
\SetKwComment{RTag}{}{}  

\title{Massively Parallel Exact Inference \\ for Hawkes Processes}

\author{%
  Ahmer~Raza \\
  Clemson University \\
  \texttt{araza@clemson.edu}
  \And
  Hudson~Smith \\
  Clemson University \\
  \texttt{dane2@clemson.edu}
}

\begin{document}

\maketitle

\begin{abstract}
  Multivariate Hawkes processes are a widely used class of self-exciting point processes, but maximum likelihood estimation naively scales as $O(N^2)$ in the number of events. The canonical linear exponential Hawkes process admits a faster $O(N)$ recurrence, but prior work evaluates this recurrence sequentially, without exploiting parallelization on modern GPUs. We show that the Hawkes process intensity can be expressed as a product of sparse transition matrices admitting a linear-time associative multiply, enabling computation via a parallel prefix scan. This yields a massively parallelizable algorithm for estimation of linear exponential Hawkes processes. Our method reduces the computational complexity to approximately $O(N/P)$ with $P$ parallel processors, and naturally yields a batching scheme to maintain constant memory usage, avoiding GPU memory constraints. Importantly, it computes the exact likelihood without any additional assumptions or approximations, preserving the simplicity and interpretability of the model. We demonstrate orders-of-magnitude speedups on simulated and real datasets, scaling to thousands of nodes and tens of millions of events, substantially beyond scales reported in prior work. We provide an open-source PyTorch library implementing our optimizations.
\end{abstract}

\section{Introduction}\label{sec:introduction}

Many natural phenomena exhibit self- and mutually-exciting behavior, whereby the occurrence of one event increases the probability of further events, either from the same source or sources that are influenced. Multivariate Hawkes processes offer a principled way to model such behavior from continuous-time event sequences, capturing mutual excitation across nodes (distinct sources of events) through the intensity function. As a flexible point process family, they have seen extensive success in various domains, particularly in finance \citep{embrechts_multivariate_2011,ait-sahalia_modeling_2015,bacry_hawkes_2015,hawkes_hawkes_2018}, seismology \citep{ogata_statistical_1988,ogata_space-time_1998,zhuang_stochastic_2002}, epidemiology \citep{unwin_using_2021,browning_simple_2021,chiang_hawkes_2022}, neuroscience \citep{truccolo_point_2016,gerhard_stability_2017,lambert_reconstructing_2018}, and social media activity \citep{zhou_learning_2013,kobayashi_tideh_2016,rizoiu_expecting_2017,rizoiu_hawkes_2017}. A key advantage of linear Hawkes processes is that the model parameters have a direct interpretation as the influence strength of one node on another, and have also been shown to encode causal dependency between nodes \citep{etesami_learning_2016,eichler_graphical_2017}.



Unfortunately, maximum likelihood estimation for linear Hawkes processes naively scales as $O(N^2)$ in the number of events, which severely limits their application to promising large-scale domains. The culprit is the computation of the intensity function at every event in the sequence, each of which requires summing contributions from all past events. This is a major reason for widespread adoption of the linear exponential Hawkes process, since its intensity function admits a sequential recurrence that allows for maximum likelihood inference in $O(N)$ time \citep{ogata_lewis_1981}. Nevertheless, scaling remains a significant issue that numerous works have sought to address, for example by exploiting sparsity in event sequences \citep{nickel_modeling_2021}, using low-rank approximations for parameters \citep{zhou_learning_2013,lemonnier_multivariate_2017,bacry_sparse_2020}, and adopting stochastic or expectation-maximization estimation algorithms \citep{veen_estimation_2008,lewis_nonparametric_2011,kirchner_estimation_2017}. However, these approaches either rely on additional assumptions about the data or introduce simplifying approximations, and can often undermine the simplicity and interpretability of the Hawkes process. Moreover, the canonical sequential recurrence method does not enable parallelization, thus leaving the potential for efficient computation on modern Graphics Processing Units (GPUs) unexploited.

We introduce a massively parallel algorithm for \textit{exact} maximum likelihood inference of linear exponential Hawkes processes, enabling modeling at formerly intractable scales. Our key observation is that the sequential intensity recurrence of the linear exponential Hawkes process admits a sparse matrix multiplication structure that can be computed in linear time. To our knowledge, this structure has not previously been identified or exploited for maximum likelihood inference of linear exponential Hawkes processes. Using this insight, we formulate our parallel algorithm, which uses the prefix scan to compute per-event intensities in approximately $O(N/P)$ time, where $N$ is the number of events and $P$ is the number of available parallel processors (often in the hundreds or thousands for modern GPUs). Our approach naturally yields a batching scheme that keeps peak memory usage constant and allows for further parallelization across GPUs, thereby circumventing GPU memory constraints. Importantly, our method computes an exact likelihood and requires \textit{no additional assumptions or approximations} to the standard linear exponential Hawkes process, thus preserving the interpretability and simplicity benefits and remaining compatible with many additional optimizations (e.g., low-rank parameterizations \citep{bacry_sparse_2020} and consideration of event sparsity \citep{nickel_modeling_2021}). We demonstrate our approach on simulated and real event sequences at previously infeasible scales, on the order of thousands of nodes and tens of millions of events. We observe orders of magnitude improvement in fitting time comparing against other implementations of the intensity computation. Finally, we release an open-source PyTorch library for fitting linear exponential Hawkes processes using our approach.\footnote{Code available at \url{https://github.com/ahmrr/HawkesTorch}.}

\section{Background}\label{sec:background}

Consider an event sequence $\mcal{H}=\left\{ (t_{i},m_{i}) \right\}_{i=1}^{N}$, where $t_{i}\in \mbb{R}_{\geq 0}$ are event timestamps ordered chronologically up to time $T$ and $m_{i}\in \left\{ 1,\dots,M \right\}$ is the node index for the $i$th event. At a particular time $t$, we only observe the event history up to $t$, denoted $\mcal{H}_{t}=\{(t_i,m_i)\in \mcal{H}: t_i < t\}$.

Let $N_p(t)$ count the events from node $p$ strictly before time $t$. Then, the conditional intensity is defined as the expected instantaneous rate of events at that node:
\begin{align*}
  \lambda_p(t\mid \mcal{H}_t) = \lim_{\Delta t \to 0^{+}} \frac{\mbb{E}\left[N_p(t+\Delta t)-N_p(t)\mid \mcal{H}_t\right]}{\Delta t}
\end{align*}

For brevity, we henceforth make the conditioning on $\mcal{H}_t$ implicit. The intensity function for a multivariate Hawkes process has the specific form $\lambda_p(t) = \mu_p(t)+\eta_p(t)$, where $\mu_p(t)$ is the base rate and $\eta_p(t)$ is the excitation rate for a particular node index $p$.

We consider the canonical linear Hawkes processes with finite mixtures of exponential kernels, characterized by the intensity function:
\begin{align}\label{eq:hawkes_excitation}
  \lambda_p(t) &= \mu_p + \sum_{j: t_{j}<t} \sum_{k=1}^{K}\alpha_{p, m_j, k} \gamma_k e^{-\gamma_k(t - t_j)}
\end{align}
$K$ is the number of decay kernels that allow modeling interactions across different timescales, each with per-kernel decay rate $\gamma_k>0$. The parameters of interest are the per-kernel interaction matrices $\mcal{A}_k\in \mbb{R}^{M\times M}_{\geq 0}$, since the elements $\alpha_{p,q,k}=(\mcal{A}_k)_{p,q}$ quantify the strength of influence of node $q$ on node $p$ at timescale $k$. Also, as $\gamma_k e^{-\gamma_k t}$ integrates to one over $t\geq 0$, we can see that $\alpha_{p,q,k}$ is the expected number of direct offspring events at node $p$ generated through kernel $k$ by one event at node $q$, which is known as the branching representation \citep{bacry_hawkes_2015}. Within the model, this interaction structure encodes dependencies between nodes in the sense of directed information \cite{etesami_learning_2016} and Granger causality \cite{eichler_graphical_2017}. For ease of exposition, we consider a constant base rate $\mu_p$, although our optimizations allow time-varying base rates, which are implemented in our library.

A Hawkes process can be fitted using maximum likelihood estimation, where the log-likelihood is:
\begin{align}\label{eq:general_hawkes_likelihood}
  \mcal{L}(\bm{\mu},\mcal{A},\bm{\gamma}) & = \sum_{i=1}^{N} \log\lambda_{m_{i}}(t_{i})- \sum_{m=1}^{M} \int_{0}^{T} \lambda_{m}(t) \, dt
\end{align}

with $\bm{\mu}=(\mu_1, \dots, \mu_M)^{\top}$, $\mcal{A} = \{\mcal{A}_1, \dots, \mcal{A}_K\}$, and $\bm{\gamma}=(\gamma_1,\dots,\gamma_K)^{\top}$. We impose a sparsity-inducing $\ell_1$ regularization on the off-diagonal elements of each $\mcal{A}_k$, since in many settings, cross-node influence is expected to be sparse while self-influence is structurally inherent \citep{bacry_sparse_2020}. The penalty is applied to elements below a certain threshold $h$, which encourages sparsity without excessively shrinking large interaction strengths. Our training objective is the per-event negative log likelihood with this penalization:
\begin{align*}
  \mrm{NLL}(\boldsymbol{\mu},\mcal{A},\boldsymbol{\gamma})
  &= -\frac{1}{N}\mcal{L}(\boldsymbol{\mu},\mcal{A},\boldsymbol{\gamma})
  + \lambda_{1}\sum_{\substack{p,q,k\\p\ne q}}
  \alpha_{p,q,k}\mb{1}\{\alpha_{p,q,k}<h\}
\end{align*}
where $\lambda_{1}\geq 0$ and $h>0$. During fitting, we enforce positivity of $\mu_p$, $\alpha_{p,q,k}$, and $\gamma_k$ using a softplus transformation: we express each parameter as $\log(1+e^z)$ for an unconstrained parameter $z\in\mbb{R}$.

With constant base rates, the Hawkes process is stable if $\rho(\sum_{k=1}^{K}\mcal{A}_k)<1$, where $\rho$ is the spectral radius \citep{bacry_hawkes_2015}. We do not enforce stability during fitting, since the likelihood on a finite observation interval is still valid and can be calculated regardless, although stability does matter when interpreting the fitted process.

In terms of efficiency, computation of the integral term in \eqref{eq:general_hawkes_likelihood} is usually straightforward and tractable (perhaps numerically if the decay kernel or base rate parameterizations prevent analytical integration). However, computing the sequence $\{\lambda_{m_{i}}(t_{i})\}_{i=1}^N$ for the log-sum term is time- and memory-intensive, naively scaling as $O(N^2MK)$. Therefore, our optimizations focus on scalable, parallelizable computation of the log-sum (i.e., the sequence $\{\eta_{m_{i}}(t_{i})\}_{i=1}^N$) and its parameter gradients.

\subsection{Sequential Recurrence for Intensity Computation}\label{sec:background_sequential_recurrence}

It is known that the Markovian nature of linear exponential Hawkes processes allows for recursively computing the likelihood \eqref{eq:general_hawkes_likelihood}, resulting in $O(N)$ scaling \citep{ogata_lewis_1981}. Due to its sequential nature, this method does not obviously allow for parallelization across the number of events $N$, making it difficult to exploit modern GPU hardware. However, it will be useful in developing our parallel method.

We define the quantity
\begin{align}\label{eq:intensity_state}
  R_p^k(t) = \sum_{j: t_{j}<t} \alpha_{p, m_j, k} e^{-\gamma_k(t - t_j)}
\end{align}
so that the excitation function \eqref{eq:hawkes_excitation} can be written as $\eta_p(t)=\sum_{k=1}^{K} \gamma_k R_p^k(t)$. We introduce the per-kernel state vector $\mb{R}_i^k = (R_1^k(t_i), \dots, R_M^k(t_i))^{\top}\in\mbb{R}^M$ for $k\in\{1,\dots,K\}$, such that our goal is to efficiently compute the sequence $\{\mb{R}_i^k\}_{i=1}^N$. Let $\Delta t_i = t_i - t_{i-1}$. The above formulation naturally leads to an affine recurrence:
\begin{align}
  \mb{R}_i^k &= e^{-\gamma_k \Delta t_i}\mb{R}_{i-1}^k + e^{-\gamma_k \Delta t_i}\bm{\alpha}_{:,m_{i-1},k} \label{eq:intensity_state_recurrence}
\end{align}
where the initial state is $\mb{R}_1^k = \mb{0}$, the $M \times 1$ zero column vector.



\subsection{Parallel Prefix Scan}\label{sub:parallel_prefix_scan}

This section provides a brief introduction to the parallel prefix scan algorithm (henceforth simply prefix scan) that we use in our approach. Given a sequence $\{a_1,a_2,\dots,a_n\}$, the prefix scan allows computing the sequence of cumulative aggregates $\{p_i\}_{i=1}^n$, each defined as $p_i = a_{i} \amul a_{i-1} \amul \dots \amul a_1$, where $\amul$ is an arbitrary associative operator that is not necessarily commutative.


Two canonical prefix scan algorithms are the Hillis-Steele scan \citep{hillis_data_1986} and the Blelloch scan \citep{blelloch_prefix_1990}. The Hillis-Steele scan has $O((n \log n) / P + \log n)$ time complexity, offering high parallelism and best performance when $n \leq P$, where $P$ is the number of available parallel processors. Conversely, the Blelloch scan has $O(n/P + \log n)$ time complexity, asymptotically performing no more operations than the sequential implementation, albeit with less parallelism. This makes it preferable when $n \gg P$, as in our setting where the number of events can reach millions while GPUs provide only thousands of cores. Moreover, the complexity reduces to $O(n/P)$ when $n/P \geq \log P$, which is the optimal speedup over the sequential implementation. Therefore, we adopt the Blelloch prefix scan.

\section{Methods}\label{sec:methods}


The crux of our approach is reformulating the affine recurrence \eqref{eq:intensity_state_recurrence} to an equivalent matrix product form using state augmentation. This transforms the affine recurrence into a linear one in a higher dimension, which will admit application of the prefix scan algorithm. Define the augmented state vector $\wt{\mb{R}}_i^k \in \mathbb{R}^{M+1}$ as:
\begin{equation}
  \wt{\mb{R}}_i^k =
  \begin{pmatrix} \mb{R}_i^k \\ 1
  \end{pmatrix}
\end{equation}
Let $I_M$ be the $M \times M$ identity matrix and $\mb{0}^{\top}$ be the $1 \times M$ zero row vector. We define the block transition matrices $\mb{M}_i^k\in\mbb{R}^{(M+1) \times (M+1)}$ as:
\begin{equation}
  \mb{M}_i^k =
  \begin{pmatrix}
    e^{-\gamma_k \Delta t_i} I_M & e^{-\gamma_k \Delta t_i} \boldsymbol{\alpha}_{:, m_{i-1}, k} \\
    \mb{0}^{\top} & 1
  \end{pmatrix}
  \label{eq:define_augmented_M}
\end{equation}
where $\mb{M}_{1}^k=I_{M+1}$. The recurrence for the augmented state vector is linear:
\begin{equation}
  \wt{\mb{R}}_i^k = \mb{M}_i^k \wt{\mb{R}}_{i-1}^k
  \label{eq:augmented_state_recurrence}
\end{equation}
Unrolling this recurrence gives the form $\wt{\mb{R}}_i^k = \mb{P}_i^k \wt{\mb{R}}_0^k$, where $\mb{P}_i^k = \mb{M}_i^k \mb{M}_{i-1}^k \dots \mb{M}_1^k$, so computing $\{\wt{\mb{R}}_i^k\}_{i=1}^N$ amounts to computing the products $\{\mb{P}_i^k\}_{i=1}^N$, which can be done using a prefix scan.\footnote{Note that applying the prefix scan directly to \eqref{eq:hawkes_excitation} by factoring out the constant $e^{-\gamma_k t_i}$ is numerically unstable: the sum of $e^{\gamma_k t_j}$ overflows while $e^{-\gamma_k t_i}$ underflows. Our formulation avoids this issue entirely by using inter-event times $\Delta t_i$.}



A key feature of this formulation is that each product of $(M+1) \times (M+1)$ transition matrices is done in $O(M)$ operations rather than the usual $O(M^3)$. To see this, consider the product of two matrices in the same format as $\mb{M}_i^k$, each defined by scalars $s_A$, $s_B$ and vectors $\mb{v}_A$, $\mb{v}_B$:
\begin{align*}
  \begin{pmatrix}
    s_A I_M & \mb{v}_A \\
    \mb{0}^{\top} & 1
  \end{pmatrix}
  \begin{pmatrix}
    s_B I_M & \mb{v}_B \\
    \mb{0}^{\top} & 1
  \end{pmatrix} &=
  \begin{pmatrix}
    s_A s_B I_M & s_A \mb{v}_B + \mb{v}_A \\
    \mb{0}^{\top} & 1
  \end{pmatrix}
\end{align*}
The resulting matrix retains the same structure, defined by a new scalar $s_{AB} = s_A s_B$ and a new vector $\mb{v}_{AB} = s_A \mb{v}_B + \mb{v}_A$. This special sparsity structure also means that each matrix can be stored using $O(M)$ memory instead of $O(M^2)$ for a dense matrix, storing only the pair $\wt{\mb{M}_i^k}=(e^{-\gamma_{k}\Delta t_i}, e^{-\gamma_{k} \Delta t_i} \boldsymbol{\alpha}_{:, m_{i-1}, k})$. This formulation lends itself to the prefix scan algorithm, where the input is the sequence of sparse transition matrices $\{\wt{\mb{M}}_i^k\}_{i=1}^N$, and the associative operation between two sparse matrices is precisely the $O(M)$ matrix multiplication described above:
\begin{align*}
  (s_A, \mb{v}_A) \amul (s_B, \mb{v}_B) = (s_A s_B, s_A \mb{v}_B + \mb{v}_A)
\end{align*}



\subsection{Efficient Parameter Gradients}

While the PyTorch automatic differentiation engine (autograd) can automatically compute the gradients of the log-likelihood, the Blelloch scan implementation leads to a complex compute graph that autograd traverses inefficiently. The upsweep and downsweep steps involve updating slices of a shared $O(MNK)$ tensor across $\lceil\log_2 N\rceil$ levels, and autograd's full gradient copies and indexing operations, through sheer memory traffic overhead, offset the Blelloch scan's arithmetic savings.

To avoid this inefficiency, we calculate the parameter gradients manually. Analogous to the intensity computation, the gradients with respect to $\alpha_{p,q,k}$ and $\gamma_k$ can be reduced to states $\{ \mb{K}_{i}^{k} \}_{i=1}^{N}$ and $\{ \mb{L}_{i}^{k} \}_{i=1}^{N}$ that admit affine recurrences, which allow us to compute them using the same prefix scan approach. The derivation of the gradients largely follows the same steps as the intensity computation, so we defer the details to Appendix~\ref{app:gradient_derivation}.

\subsection{Time Complexity Analysis and Batching}
\label{sub:complexity_batching}

We discuss the general time complexity of our method, in terms of both memory and compute; specifically, we examine the complexity of the three prefix scans required across the forward and backward passes. Along the kernel dimension $K$, the cost of each associative matrix multiplication is $O(KM)$, meaning that the prefix scan operates in time $O(KMN/P)$ for $P$ parallel processors. For each prefix scan, the peak memory required for storing the compressed prefix matrices is $O(KMN)$. This may be prohibitive when fitting very large event sequences, e.g. $M\gtrsim 10^3$ and $N\gtrsim 10^6$. We thus outline a method for calculating gradients in a batched manner, so as to guarantee constant peak memory usage.

Consider partitioning the sequence $\mcal{H}$ into $B$ batches. Let $\mcal{I}=\{1,\dots,N\}$ denote the index set for all events, and define $N_{b}$ as the size of batch $b$. Let
$$
\mcal{I}_{b}=\left\{ s_b+1, s_b+2, \dots, s_b+N_{b} \right\}
$$
denote the index set for a batch $b=1,\dots,B$, where $s_b=\sum_{i=1}^{b-1} N_i$ is the starting offset of batch $b$.

In the forward pass, for each batch $b$, we can independently compute the per-batch states $\{ \mb{R}_{i}^{k} \}_{i\in \mcal{I}_{b}}$ using the final state of the previous batch $\mb{R}_{s_b}^{k}$, since for any index $l < i$:
\begin{align*}
  \wt{\mb{R}}_{i}^{k} = \mb{P}_{i}^{k} \wt{\mb{R}}_{0}^{k} = \left[\mb{M}_{i}^{k} \dots \mb{M}_{l+1}^{k} \right] \wt{\mb{R}}_{l}^{k}
\end{align*}
We can also compute the states needed for gradient calculations in a similar batched fashion; see Appendix~\ref{app:gradient_derivation} for the details. A key benefit of this batching approach is that each batch depends on the previous only through a single state. Therefore, the products for each batch can be computed independently, perhaps parallelized across different GPUs. Importantly, this batching scheme does not approximate the likelihood or gradients. They are computed exactly for the entire event sequence, and batching is used purely to control memory consumption.

To illustrate the approach, Algorithm~\ref{alg:training_sequence} presents the full training procedure for the unbatched case. For clarity, we treat the kernel dimension $k$ as independent and vectorized, so no explicit loop over $k$ is included. The batched version of the procedure is a straightforward extension, and is given in Appendix~\ref{app:batched_algorithm}.

\begin{algorithm}
  \caption{Massively parallelizable training of multivariate Hawkes processes}
  \label{alg:training_sequence}

  \SetKwBlock{Forward}{\textnormal{\textsc{Forward Pass}}}{}
  \SetKwBlock{Backward}{\textnormal{\textsc{Backward Pass}}}{}

  \textbf{define} $\mathrm{PrefixScan}_{\amul}(\,\cdot\,)$ \textbf{from} Section~\ref{sub:parallel_prefix_scan}\;

  \For{\textnormal{each epoch} $\ell$}{

    \Forward{
      $\left\{(\ast, \mb{R}_i^k)\right\}_{i\in\mcal{I}} \gets \mathrm{PrefixScan}_{\amul}\left(\left\{(e^{-\gamma_k \Delta t_i}, e^{-\gamma_k \Delta t_i}\bm{\alpha}_{:,m_{i-1},k})\right\}_{i\in\mcal{I}}\right)$\;

      $\{(\ast, \mb{L}_i^k)\}_{i\in\mcal{I}} \gets \mathrm{PrefixScan}_{\amul}\left(\left\{(e^{-\gamma_k \Delta t_i}, \Delta t_i\mb{R}_i^k)\right\}_{i\in\mcal{I}}\right)$\;

      $\lambda_{m_i}(t_i) \gets \mu_{m_i} + \sum_{k=1}^{K} \gamma_k R_{m_i}^k(t_i)$, $\forall i\in\mcal{I}$\;

      $\mcal{L} \gets \sum_{i\in\mcal{I}} \log \lambda_{m_i}(t_i) - \sum_{p=1}^{M}\int_0^T\lambda_p(t)\,dt$\;

      $D_i^k \gets R_{m_i}^k(t_i)-\gamma_k L_{m_i}^k(t_i)$, $\forall i\in\mcal{I}$\;

      \textbf{store} $\{\lambda_{m_i}(t_i)\}_{i\in\mcal{I}}$, $\{D_i^k\}_{i\in\mcal{I}}$ for backward pass\;
    }

    \Backward{
      $\{(\ast, \mb{K}_i^k)\}_{i\in\mcal{I}} \gets \mathrm{PrefixScan}_{\amul}\left(\left\{(e^{-\gamma_k \Delta t_i}, e^{-\gamma_k \Delta t_i}\mb{e}_{m_{i-1}})\right\}_{i\in\mcal{I}}\right)$\;

      $\nabla\mu_p \gets$ using Equation~\eqref{eq:grad_mu} and $\{\lambda_{m_i}(t_i)\}_{i\in\mcal{I}}$\;

      $\nabla\alpha_{p,q,k} \gets$ using Equation~\eqref{eq:grad_alpha} and $\{\mb{K}_i^k\}_{i\in\mcal{I}}$\;

      $\nabla\gamma_{k} \gets$ using Equation~\eqref{eq:grad_gamma} and $\{D_i^k\}_{i\in\mcal{I}}$\;
    }

    $\mu_p, \alpha_{p,q,k}, \gamma_k \gets \mathrm{OptimizerStep}(\mu_p, \alpha_{p,q,k}, \gamma_k, \nabla\mu_p, \nabla\alpha_{p,q,k}, \nabla\gamma_k, \mathrm{lr})$\;
  }
\end{algorithm}

\section{Results}\label{sec:results}

We evaluate our method on both simulated and real-world datasets to demonstrate its scalability and the typical conditions under which it outperforms other implementations. Because our algorithm does not alter the likelihood or gradients, we focus our empirical evaluation on computational efficiency, and include a single numerical example confirming the implementation preserves standard Hawkes process behavior. Using synthetic data, we study scaling behavior and conduct an ablation study, comparing against different implementations of the intensity. We then demonstrate performance on real datasets that were formerly intractable with standard approaches, namely MemeTracker and Chicago crime data, to show both the scalability and interpretability of our approach.

All experiments were run primarily on an Nvidia A100 80GB GPU. When comparing against CPU implementations, we used 64 cores of the fastest available CPU on our cluster, the Intel Xeon Platinum 8580. We use single-precision floating-point arithmetic (\texttt{float32}) and, unless otherwise specified, the Adam optimizer with learning rate $0.1$, penalty weight $\lambda_1=0.01$, and hinge point $h=0.05$. For fits to real data, we set an initial batch size of $2^{18}$ events, and halve it if needed to fit into memory.

\subsection{Scaling on Synthetic Data}

We now study the scaling of our algorithm on massive simulated event sequences. Here, we use $\gamma=1.5$ and simulate from a hub-and-spoke influence network, where a single row of the interaction matrix $\mcal{A}$ is set to $0.1$, and the base rate $\mu_p$ of the single hub node is $0.1$. We consider event sequences of length equal to powers of two, since the prefix scan internally pads shorter sequences to the next power of two. The sequence lengths range from $N=2^{10}$ to $N=2^{26}$ (i.e., $N\sim10^3$ to $N\sim 6.7 \times 10^7$), and the number of nodes ranges from $M = 125$ to $2000$, doubling at each step. We use $K=3$ exponential decay kernels with parameterized decay rates. All runs are performed on the GPU, and each sequence was fitted for $1000$ epochs, with a maximum allowed wall time of $72$ hours.

\begin{figure}[h]
  \centering
  \includegraphics[width=1\linewidth]{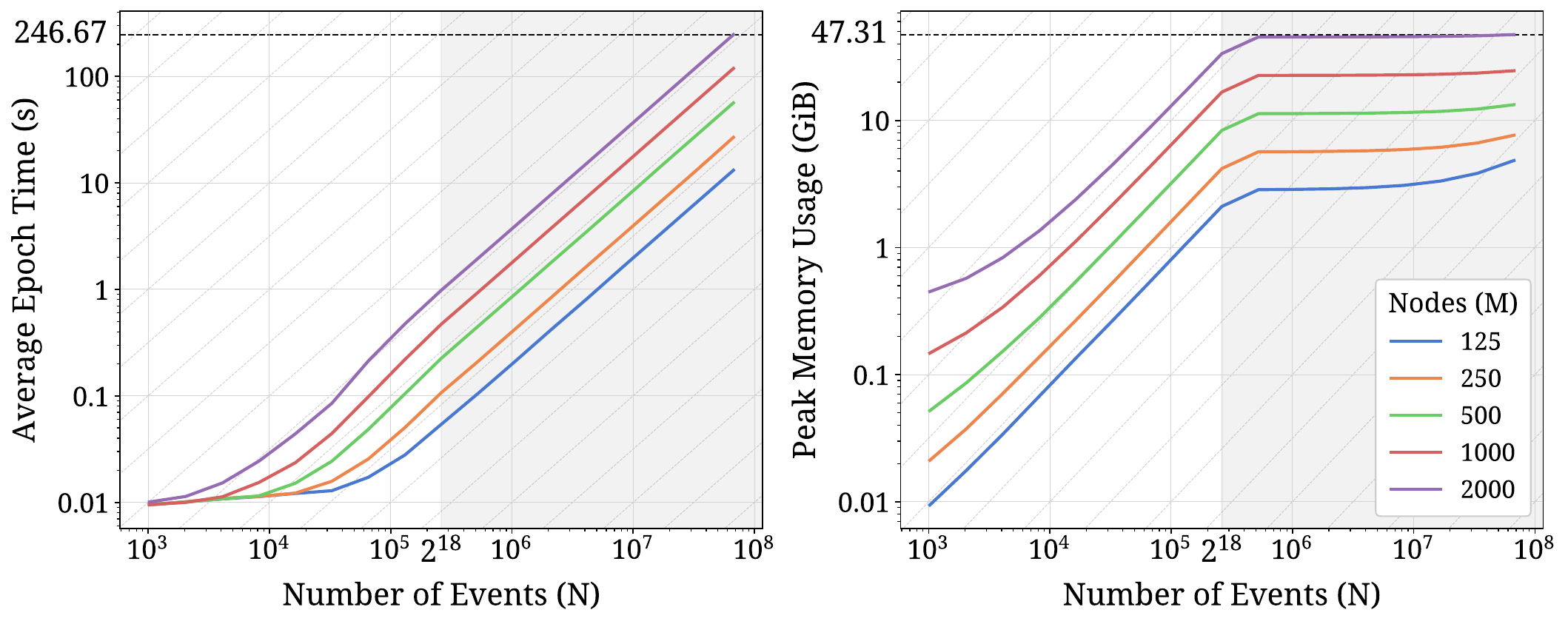}
  \caption{Scaling of our method across various sequence lengths, showing average epoch time (left) and peak memory usage (right). The gray region indicates where batching is enabled to maintain constant memory (with batch size $2^{19}=\num{524288}$). Dashed black lines indicate the maximum epoch time and memory usage across all runs, respectively. Diagonal gridlines with unit power (slope 1 on log-log plot) are shown for reference.}
  \label{fig:performance_scaling}
\end{figure}

Figure~\ref{fig:performance_scaling} shows time and memory scaling across the simulated sequences. As expected from theoretical scaling, the epoch time and memory grow linearly with the number of events (and the number of nodes) for larger event sequences. In the small $N$ regime, scaling is sublinear due to vacant parallel cores on the GPU, and then transitions to linear once all the cores saturate. Once batching begins, the memory usage peaks at 47.31 GiB, which can be adjusted via the batch size to match the target hardware. The performance curves are remarkably stable due to the prefix scan: each doubling of $N$ causes a consistent doubling in the average epoch time, which peaks at $246.67$ seconds.

\subsection{Comparison to Other Likelihood Computation Methods}\label{sec:results_comparison}

On the same synthetic data, we conduct an ablation study comparing different implementations of the likelihood computation. The naive baseline implementation, which scales quadratically in the number of events, involves directly computing the sum over previous events for each intensity $\lambda_{m_i}(t_i)$. The sequential implementation uses the recurrence formulas \eqref{eq:intensity_state_recurrence}, \eqref{eq:alpha_grad_recurrence}, and \eqref{eq:gamma_grad_recurrence}, and is tested on both the GPU and CPU. The CPU version is given $64$ CPU cores, although there is no parallelization and thus negligible benefit from providing more cores. Finally, we test our prefix scan-based algorithm both using PyTorch's default autograd gradients and our manual gradient computation. All runs are fitted for $1000$ epochs and limited to $24$ hours.

\begin{figure}[t]
  \centering
  \includegraphics[width=1\linewidth]{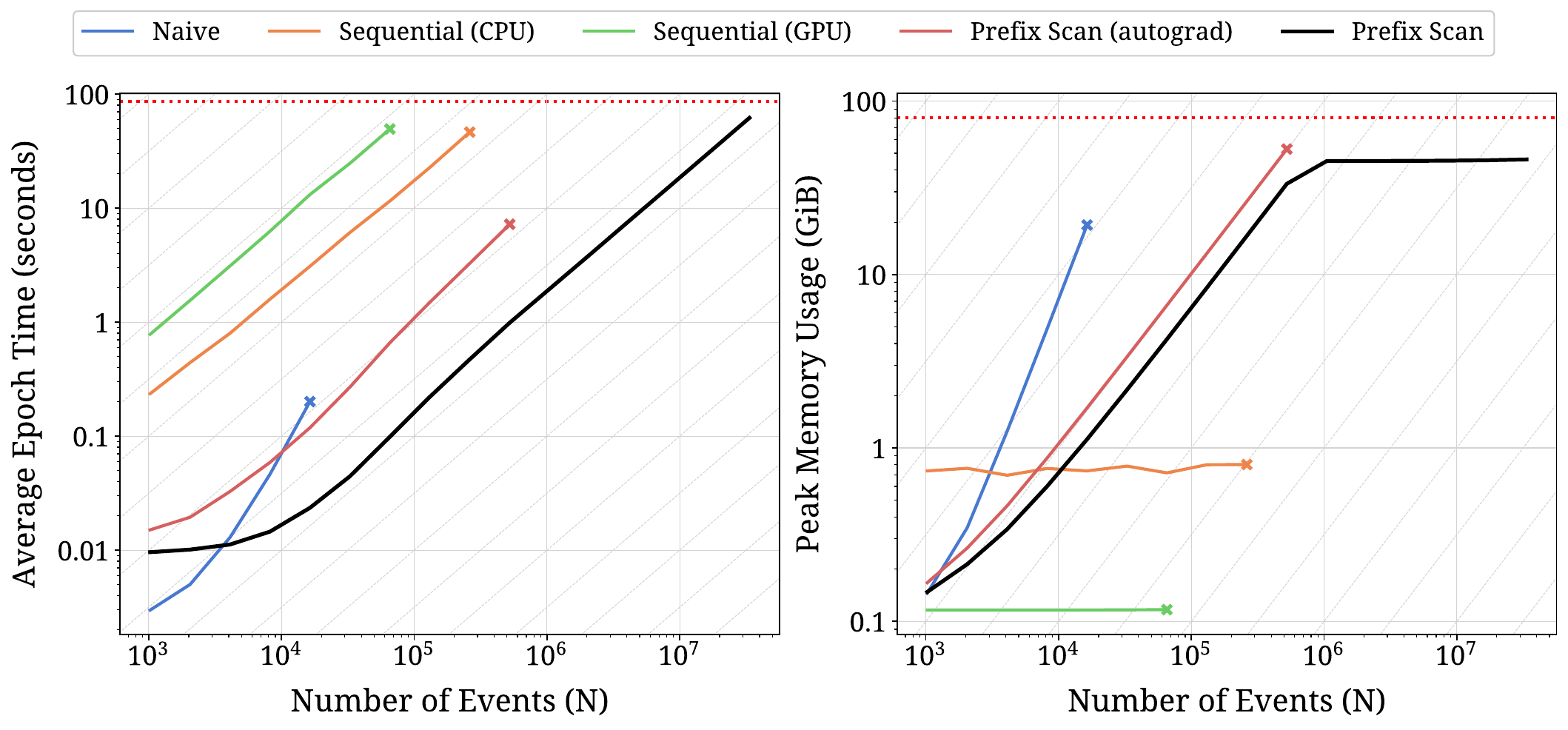}
  \caption{Comparison between our method (black) and other implementations of the intensity computation. The red dotted lines indicate the 24-hour walltime limit (86.4 seconds per epoch for 1000 epochs) and the maximum available memory (80 GiB), respectively. Crosses indicate that the next run either ran out of memory or exceeded 24 hours.}
  \label{fig:performance_comparison}
\end{figure}

Figure~\ref{fig:performance_comparison} presents the per-epoch time and peak memory usage of each ablation configuration. The naive method runs in quadratic time and is heavily memory-bound, terminating within $10^5$ events. The sequential methods (both on the GPU and CPU) have the expected linear runtime and constant peak memory usage, but the constant factor is very large and, thus, the sequential implementations terminate early. The CPU implementation outperforms the equivalent GPU version because the sequential intensity computation is not parallelizable, so running on the GPU incurs significant overhead and suffers from weak single-core performance. The prefix scan implementation with PyTorch autograd gradients scales significantly better than the other methods, but does not provide an easy way to do batching and, therefore, is memory-limited. Finally, the full implementation using the prefix scan is, by far, the most performance efficient, being orders of magnitude faster than the alternative implementations and scalable to arbitrary $N$ due to the constant memory usage.

In Appendix~\ref{app:comparison}, we provide an extended version of this study. We run our approach (for both autograd and manually-computed gradients) on the CPU as well; as to be expected, it is slower than the GPU version, but is still substantially faster than the other implementations. Also, we compare runtimes of the Blelloch versus Hillis-Steele scan implementations, finding that our choice of the Blelloch scan is indeed most efficient for the large $N$ regime we explore. Finally, we extend all experiments to the low $N$ regime (down to $N=8$), which more clearly highlights the cross-over points for all methods.


\subsection{Parameter Recovery Study}

While estimation accuracy is not the primary focus of this work, we include a synthetic experiment to verify that our proposed algorithm preserves standard Hawkes parameter recovery behavior. Across $30$ random initializations, we simulate event sequences from a single-kernel multivariate Hawkes process using Ogata’s thinning algorithm \citep{ogata_lewis_1981}. For all dimensions $p$, the base rate is fixed to $\mu_p = 0.001$, and a single exponential decay kernel with rate $\gamma = 1.0$ is used. We vary the number of dimensions $M \in \{50, 100, 250, 500, 1000\}$ while fixing the total number of events to $N=\num{1000000}$. The interaction matrix $\mathcal{A}$ is generated from a sparse directed scale-free graph \citep{bollobas_directed_2003,barabasi_scale-free_2009}. The penalty weight and hinge point are set to $\lambda_1 = 0.1$ and $h = 0.05$, respectively. The synthetic event sequences are then fitted using a single-kernel Hawkes model with the Adam optimizer, with $1000$ epochs and a learning rate of $0.05$.

Table~\ref{tab:parameter_recovery_results} reports the per-event log-likelihood and relative $\ell_2$ error of the estimated parameters with respect to their true values, which are then averaged across the 30 seeds. All values in the table have corresponding $t$-confidence intervals; note that they are not confidence intervals for the true parameter. Across all settings, the decay rate $\gamma$ is consistently estimated with less than $2\%$ relative error. When the number of events per interaction parameter $N/M^2$ is moderate (that is, $M \leq 250$ so $N/M^2 \geq 16$), both the base rate $\mu$ and interaction matrix $\mcal{A}$ are recovered with error below $10\%$. However, more underdetermined settings ($M=1000$ and $500$, so $N/M^2=1$ and $4$) exhibit much higher relative errors; for example, the error reaches $62.53\%$ when $M=1000$. In these cases, most interactions are weakly informed by the data, and the $\ell_1$ penalty further shrinks weakly supported elements of $\mathcal{A}$ toward zero. Nonetheless, the parameters remain well-recovered in most scenarios, and the model still captures the global dynamics of the process.

\begin{table}[h]
  \centering

  \caption{Synthetic parameter recovery results. Per-event log-likelihood is reported. Brackets indicate the 95\% confidence interval. The displayed relative $\ell_2$ errors are scaled by $10^{-2}$ and so can be interpreted as percentages.}
  \label{tab:parameter_recovery_results}

  \medskip


  \begin{tabular}{c r@{\hspace{4pt}}l r@{\hspace{4pt}}l
    r@{\hspace{4pt}}l r@{\hspace{4pt}}l}
    \toprule
    & \multicolumn{2}{c}{}
    & \multicolumn{6}{c}{\textbf{Relative $\ell_2$ Error}} \\
    \cmidrule(lr){4-9}
    \textbf{Nodes}
    & \multicolumn{2}{c}{\textbf{Log-Lik.}}
    & \multicolumn{2}{c}{$\boldsymbol{\mu}\;(\times 10^{-2})$}
    & \multicolumn{2}{c}{$\boldsymbol{\mathcal{A}}\;(\times 10^{-2})$}
    & \multicolumn{2}{c}{$\boldsymbol{\gamma}\;(\times 10^{-2})$} \\
    \midrule
    50
    & $-2.836$ & \raisebox{0.8pt}{\footnotesize{$[-2.839,\ -2.832]$}}
    & $2.99$ & \raisebox{0.8pt}{\footnotesize{$[2.91,\ 3.07]$}}
    & $1.95$ & \raisebox{0.8pt}{\footnotesize{$[1.88,\ 2.01]$}}
    & $1.34$ & \raisebox{0.8pt}{\footnotesize{$[1.03,\ 1.64]$}} \\[2pt]
    100
    & $-3.117$ & \raisebox{0.8pt}{\footnotesize{$[-3.120,\ -3.113]$}}
    & $6.07$ & \raisebox{0.8pt}{\footnotesize{$[6.01,\ 6.13]$}}
    & $2.93$ & \raisebox{0.8pt}{\footnotesize{$[2.85,\ 3.01]$}}
    & $0.55$ & \raisebox{0.8pt}{\footnotesize{$[0.39,\ 0.71]$}} \\[2pt]
    250
    & $-4.548$ & \raisebox{0.8pt}{\footnotesize{$[-4.552,\ -4.544]$}}
    & $9.29$ & \raisebox{0.8pt}{\footnotesize{$[9.24,\ 9.33]$}}
    & $4.41$ & \raisebox{0.8pt}{\footnotesize{$[4.34,\ 4.48]$}}
    & $0.91$ & \raisebox{0.8pt}{\footnotesize{$[0.71,\ 1.11]$}} \\[2pt]
    500
    & $-4.693$ & \raisebox{0.8pt}{\footnotesize{$[-4.699,\ -4.686]$}}
    & $18.85$ & \raisebox{0.8pt}{\footnotesize{$[18.80,\ 18.89]$}}
    & $10.20$ & \raisebox{0.8pt}{\footnotesize{$[9.71,\ 10.68]$}}
    & $1.62$ & \raisebox{0.8pt}{\footnotesize{$[1.47,\ 1.77]$}} \\[2pt]
    1000
    & $-5.785$ & \raisebox{0.8pt}{\footnotesize{$[-5.792,\ -5.777]$}}
    & $39.42$ & \raisebox{0.8pt}{\footnotesize{$[39.23,\ 39.62]$}}
    & $62.53$ & \raisebox{0.8pt}{\footnotesize{$[62.40,\ 62.66]$}}
    & $0.64$ & \raisebox{0.8pt}{\footnotesize{$[0.54,\ 0.73]$}} \\
    \bottomrule
  \end{tabular}

\end{table}

\subsection{Confirming Exactness of the Algorithm}

While our prefix scan formulation with batching is mathematically equivalent to the standard Hawkes process likelihood, the prefix scan reorders operations, which can introduce small numerical differences due to non-associativity of floating-point arithmetic. To confirm that our implementation is correct and exact in practice, we compare our algorithm's computed intensity and parameter gradients against sequential and naive implementations on synthetic data.

We set $M=10$ and $K=3$, and for each of $5$ random seeds, simulate event sequences of lengths $N\in\{10^2, 10^3, 10^4, 10^5, 10^6\}$. To each sequence from each seed, we fit a Hawkes process for $10$ epochs using our approach, and at each epoch also calculate the intensity and gradients using the naive and sequential implementations. Finally, we compute the index-wise relative errors between all three methods, and report the maximum relative error across all indices, epochs, and seeds. We defer the full results to Appendix~\ref{app:correctness}, but across all sequence lengths, the maximum relative error compared to the sequential implementation is less than $10^{-5}$ for the intensity and less than $10^{-4}$ for the gradients. This indicates that our implementation is also exact in practice, up to negligible numerical differences due to floating-point arithmetic.

\subsection{MemeTracker Data}

%
%

To examine the runtime performance of our approach on real information cascades, we use the MemeTracker dataset \cite{leskovec_meme-tracking_2009} as curated by \cite{gomez_rodriguez_structure_2013}. The full dataset consists of $\num{34547727}$ unique memes posted on $5000$ of the most active websites between March 2011 and February 2012. Memes are grouped into 178 categories based on shared keywords, where each event corresponds to a meme posted on a specific website at a particular time. We fit models to ten representative subsets spanning approximately $6.4\times 10^4$ to $3.2\times 10^7$ events, using $K=1$ decay kernel and $1000$ training epochs.

Results for all subsets are reported in Table~\ref{tab:memetracker_results}. For comparison, we include average per-epoch runtimes on four subsets reported by \cite{nickel_modeling_2021}, whose method uses a low-rank parameterization and exploits sparsity in event sequences. Our approach targets likelihood computation, so this comparison is intended to highlight the potential gains from likelihood speedups alone, without supplementary optimizations.

Across the four shared subsets, our per-epoch runtimes are approximately an order of magnitude lower than the reported Lazy MHP timings. The gains are smaller for shorter event sequences, as expected since the prefix scan offers more speedup as sequence length grows. Moreover, our approach successfully fits substantially larger subsets, up to approximately $3.2\times 10^7$ events. To the best of our knowledge, this is the first time the standard multivariate Hawkes process, fit via exact maximum likelihood estimation, has been applied at this scale.

\begin{table}[h]
  \centering
  \caption{Summary of MemeTracker subsets and fitted Hawkes process models. Per-event log-likelihood is reported. Lazy MHP results are obtained from \cite{nickel_modeling_2021} and are omitted for subsets not considered in the paper. This is not a controlled hardware comparison between our method and Lazy MHP; we simply intend to demonstrate the potential gains from likelihood speedups alone.}
  \label{tab:memetracker_results}

  \medskip

  \begin{tabular}{rrrccc}
    \toprule
    \multicolumn{4}{c}{} & \multicolumn{2}{c}{\textbf{Avg. Epoch Time}} \\
    \cmidrule(lr){5-6}
    \textbf{Subset} & \textbf{Nodes} & \textbf{Events} & \textbf{Log-Lik.} & \textbf{Ours} & \textbf{Lazy MHP} \\
    \midrule
    Bail Out & 835 & 64,138 & 3.612 & 0.06\,s & 0.47\,s \\
    Miami Heat & 1,173 & 133,451 & 1.680 & 0.21\,s & 1.60\,s \\
    Amy Winehouse & 1,561 & 226,247 & 2.043 & 0.34\,s & 3.30\,s \\
    Arab Spring & 1,377 & 400,199 & 2.796 & 0.46\,s & 4.60\,s \\
    Greece & 1,925 & 1,077,125 & 2.280 & 1.24\,s & \textemdash \\
    Libya & 2,003 & 2,486,030 & 2.679 & 2.97\,s & \textemdash \\
    Crisis & 2,266 & 5,085,759 & 2.495 & 6.94\,s & \textemdash \\
    Leader & 2,344 & 10,604,129 & 2.674 & 14.91\,s & \textemdash \\
    North Korea... & 2,451 & 16,408,957 & 2.747 & 23.73\,s & \textemdash \\
    Prince William... & 2,494 & 32,687,780 & 2.827 & 48.33\,s & \textemdash \\
    \bottomrule
  \end{tabular}
\end{table}

%

\subsection{Chicago Crime Data}

Finally, to demonstrate an analysis at a previously infeasible scale, we fit multivariate Hawkes processes to incident-level crime data from Chicago. Crimes exhibit both self excitation within an area and cross excitation between neighboring areas, making Hawkes processes well-suited to model crime diffusion \citep{mohler_self-exciting_2011,mohler_randomized_2015,zhuang_semiparametric_2019}.

We analyze a snapshot of the Chicago Police Department dataset of crimes since 2001 \citep{chicago_police_department_crimes_2025}, downloaded on November 14, 2025 and containing \num{7834628} valid crime entries. Since crime spreads differently depending on the type, we fit models to all crimes and to five separate subsets containing theft, battery, assault, burglary, or robbery. The nodes of our models correspond to the $22$ police districts where crimes occurred. We use $K=3$ decay kernels and fit each model for $1000$ epochs.

\begin{figure}[h]
  \centering

  \begin{subfigure}{1\textwidth}
    \centering
    \includegraphics[width=0.98\linewidth]{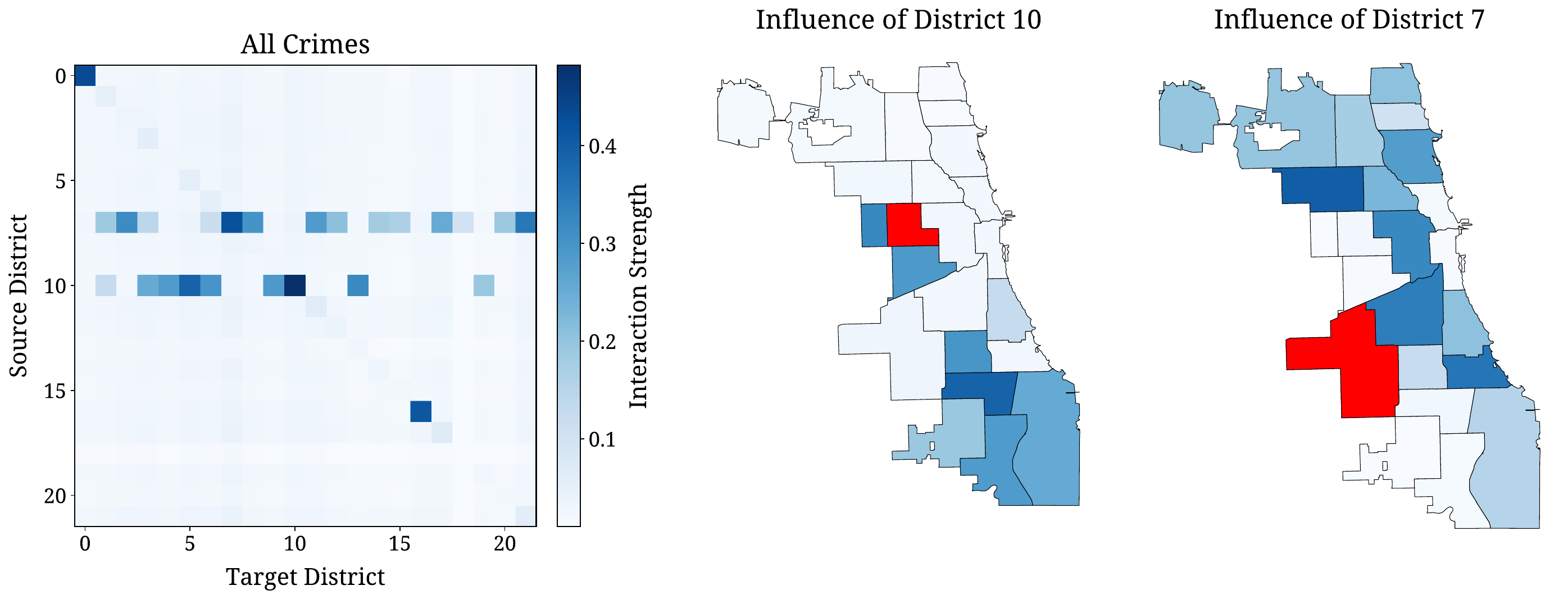}
    \phantomsubcaption\label{fig:all_crimes_influence}
  \end{subfigure}
  \begin{subfigure}{1\textwidth}
    \centering
    \includegraphics[width=0.99\linewidth]{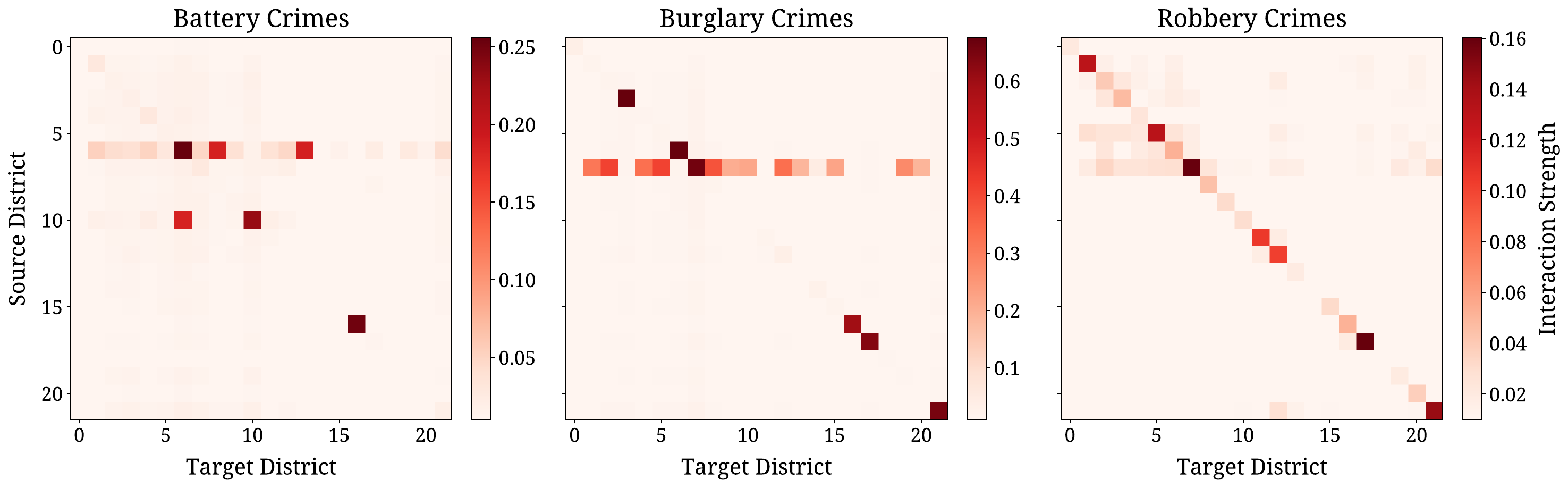}
    \phantomsubcaption\label{fig:selected_crimes_influence}
  \end{subfigure}

  \caption{Top row: the interaction matrix for all crime types learned from Chicago crime data, and choropleth maps showing influence of particularly influential districts (District 7 and District 10). Bottom row: three interaction matrices for selected crime types.}
\end{figure}

A summary of the data and fitted models is given in Table~\ref{tab:crime_data_results}, while Figures~\ref{fig:all_crimes_influence} and \ref{fig:selected_crimes_influence} show district influence maps and selected interaction matrices. When considering all crimes, there is a clear hub-and-spoke structure, whereby two districts influence crime in many others. On the other hand, individual crime types show different behaviors; battery influence is sparse, while burglary and robbery exhibit strong self-influence. These learned spatial patterns are broadly consistent with district-level crime patterns reported by the Chicago Police Department \citep{chicago_police_department_2024_2024}. Across all fitted models, the maximum spectral radius is $\rho(\sum_{k=1}^{K}\mcal{A}_k)=0.89<1$, which satisfies the stability condition.

\begin{table}[h]
  \centering

  \caption{Summary of Chicago crime data and the fitted Hawkes models with $M=22$ nodes. Per-event log-likelihood is reported.}
  \label{tab:crime_data_results}

  \medskip

  \begin{tabular}{rrcc}
    \toprule
    \textbf{Crime Types} & \textbf{Events ($N$)} & \textbf{Avg. Epoch Time} & \textbf{Log-Likelihood} \\
    \midrule
    All & 7,834,628 & 1.012\,s & 1.012 \\
    Selected & 4,761,820 & 0.605\,s & 0.428 \\
    Theft & 2,071,626 & 0.296\,s & -0.189 \\
    Battery & 1,420,739 & 0.200\,s & -1.565 \\
    Assault & 563,535 & 0.082\,s & -2.720 \\
    Burglary & 414,073 & 0.072\,s & -2.454 \\
    Robbery & 291,847 & 0.051\,s & -3.498 \\
    \bottomrule
  \end{tabular}
\end{table}




\section{Related Work}\label{sec:related_work}

We briefly discuss the two lines of research that our work brings together, namely scalable/parallel inference for Hawkes processes and using parallel scans for inference in sequential models.

\textbf{Scalable Hawkes process inference.} As discussed in Section~\ref{sec:background_sequential_recurrence}, recursively evaluating the likelihood for exponential Hawkes processes is a classical method \citep{ozaki_maximum_1979}. In the same vein of scalable inference, there have been subsequent works that exploit sparsity in event sequences \citep{nickel_modeling_2021} and sparse or low-rank interaction structures \citep{bacry_sparse_2020}. Several works have also explored hardware acceleration, including FPGA implementations for multivariate exponential models \citep{guo_accelerating_2014} and GPU likelihood computation for Bayesian spatiotemporal models \citep{holbrook_scalable_2021,ko_scaling_2024}. Our approach is distinct from these hardware approaches, as it fundamentally addresses the sequential nature of the exponential recurrence, enabling parallel computation of intensities and analytic parameter gradients across events. The model and likelihood are unchanged, and hence remain compatible with these optimizations as well.

\textbf{Parallel scans for inference.} Prefix scans are a well-established primitive for parallelizing recurrences \citep{blelloch_prefix_1990}, and so have been applied to many types of sequence models. This includes linear RNNs \citep{martin_parallelizing_2018}, state space sequence models such as S5 and Mamba \citep{smith_simplified_2023,gu_mamba_2024}, and Bayesian filtering and smoothing \citep{sarkka_temporal_2021}. A recent work even aggregates many such optimizations into a broader framework for prefix-scannable sequence models \citep{yau_sequential_2026}. In these settings, efficiently computing the associative operation relies on some explicitly imposed structure, for example, a sparse or low-rank matrix. For exponential Hawkes processes, we note that the sparse scalar decay structure is intrinsic to the model, which allows the $O(M)$ operation without any additional restrictions on the interaction matrices.

\section{Conclusion}\label{sec:conclusion}

%
%

We presented a massively parallelizable algorithm for exact maximum likelihood estimation of linear exponential Hawkes processes. We believe this work opens the door to applications in promising new domains where the scale of event sequences has thus far prevented exact Hawkes process inference, and to this end provide a modular, open-source PyTorch library that implements our approach.

A few limitations remain worth noting. Our method relies on parallelism to achieve its gains, so without a GPU or many CPU cores, or for small event sequences, the prefix scan offers little advantage. Moreover, our method applies to linear Hawkes processes with finite mixtures of exponential kernels, and does not directly extend to the case of arbitrary kernel families.

\begin{ack}
  We gratefully acknowledge funding from the Media Forensics Hub at Clemson University and the use of the Palmetto supercomputer provided through this support. There are no competing interests to declare.
\end{ack}

\bibliography{bibliography}


\newpage
\appendix
\section{Parameter Gradient Formulas}
\label{app:gradient_derivation}

From the log-likelihood definition \eqref{eq:general_hawkes_likelihood}, we can write its derivatives with respect to the relevant parameters as:
\begin{align}
  \frac{\partial\mcal{L}}{\partial\mu_p} &= \sum_{\substack{i:m_i=p}} \frac{1}{\lambda_{m_i}(t_i)} - T \label{eq:grad_mu} \\
  \frac{\partial\mcal{L}}{\partial\alpha_{p,q,k}} &= \sum_{i:m_i=p} \frac{\gamma_k K_{q}^k(t_i)}{\lambda_{m_i}(t_i)} - \sum_{i:m_i=q} \left(1-e^{-\gamma_k(T-t_i)}\right) \label{eq:grad_alpha} \\
  \frac{\partial\mcal{L}}{\partial\gamma_k} &= \sum_{i=1}^{N}\left[\frac{D_i^k}{\lambda_{m_i}(t_i)} \right] - \sum_{m=1}^{M}\sum_{i=1}^{N} \alpha_{m,m_i,k} (T-t_i) e^{-\gamma_k(T-t_i)} \label{eq:grad_gamma}
\end{align}
where each component of the per-kernel gradient state vectors $\mb{K}_i^k = (K_1^k(t_i), \dots, K_M^k(t_i))^{\top}\in\mbb{R}^M$ and $\mb{L}_i^k=(L_1^k(t_i), \dots, L_M^k(t_i))^{\top}\in\mbb{R}^M$ is:
\begin{align*}
  K_{q}^k(t) &= \sum_{j:t_j < t} \delta_{m_j,q} e^{-\gamma_k(t - t_j)} \\
  L_{p}^k(t) &= \sum_{j:t_j<t} \alpha_{p,m_j,k} (t-t_j) e^{-\gamma_k(t - t_j)}
\end{align*}
and for the $\gamma_k$ gradient, the component that depends on the prefix scan results is:
\begin{align*}
  D_i^k=\frac{\partial}{\partial\gamma_k}\lambda_{m_i}(t_i)=R_{m_i}^k(t_i)-\gamma_k L_{m_i}^k(t_i)
\end{align*}

Using this form of the $\gamma_k$ gradient, we are able to work with the inter-event times directly, thus avoiding the issue of subtracting large terms that depend on the absolute time.

Overall, this formulation yields affine recurrences for the state vectors:
\begin{align}
  \mathbf{K}_i^k &= e^{-\gamma_k \Delta t_i}\mathbf{K}_{i-1}^k + e^{-\gamma_k \Delta t_i} \mathbf{e}_{m_{i - 1}} \label{eq:alpha_grad_recurrence} \\
  \mathbf{L}_i^k &= e^{-\gamma_k\Delta t_i}\mathbf{L}_{i-1}^k + \Delta t_i\mathbf{R}_i^k \label{eq:gamma_grad_recurrence}
\end{align}
Both states start at zero at the first event. Equivalently, using state augmentation:
\begin{align*}
  \wt{\mb{K}}_i^k =
  \begin{pmatrix}
    \mathbf{K}_i^k \\ 1
  \end{pmatrix} &=
  \begin{pmatrix}
    e^{-\gamma_k \Delta t_i} I_M & e^{-\gamma_k \Delta t_i}\mathbf{e}_{m_{i-1}} \\
    \mathbf{0}^{\top} & 1
  \end{pmatrix}
  \begin{pmatrix}
    \mathbf{K}_{i - 1}^k \\ 1
  \end{pmatrix} \\
  \wt{\mb{L}}_i^k =
  \begin{pmatrix}
    \mathbf{L}_i^k \\ 1
  \end{pmatrix} &=
  \begin{pmatrix}
    e^{-\gamma_k \Delta t_i} I_M & \Delta t_i\mathbf{R}_i^k \\
    \mathbf{0}^{\top} & 1
  \end{pmatrix}
  \begin{pmatrix}
    \mathbf{L}_{i - 1}^k \\ 1
  \end{pmatrix}
\end{align*}
where $\mathbf{e}_i\in\{0,1\}^M$ denotes the vector of $0$'s with a single $1$ at the $i$th position. The state $K_q^k(t_i)$ is shared across destination nodes $p$, so a single sequence $\{\mb{K}_i^k\}_{i=1}^N$ suffices for all interaction gradients.

Similar to the batched calculation of $\{\mb{R}_i^k\}_{i=1}^N$ described in Section~\ref{sub:complexity_batching}, we can compute the states $\{\mb{K}_i^k\}_{i=1}^N$ and $\{\mb{L}_i^k\}_{i=1}^N$ (without any approximation) using batching. Each batch's prefix scans are applied to the final states $\mb{K}_{s_b}^k$ and $\mb{L}_{s_b}^k$ of the previous batch, preserving contributions from all earlier events. The forward pass computes $\mb{R}$ and then $\mb{L}$ within each batch, saving only $D_i^k$ and $\lambda_{m_i}(t_i)$ for backward. The backward pass scans $\mb{K}$ and accumulates the gradient sums in \eqref{eq:grad_mu}--\eqref{eq:grad_gamma}. All the full batch states are discarded, except for the final states which are used as initial states for the next batch.

\newpage
\section{Batched Training Algorithm}
\label{app:batched_algorithm}

In this section, we present the extension of Algorithm~\ref{alg:training_sequence} to the batched case; that is, the forward and backward passes for training multivariate Hawkes processes using our approach, split across batches.

Note that in order to achieve constant memory usage, each batch's forward pass must immediately be followed by its backward pass. Only the marked-node derivatives $D_i^k$ and intensities are retained across the two passes; all the full states per batch are discarded after copying the final states needed by the next batch. The parameter gradients are accumulated across batches, and each batch's stored intensities and derivatives are discarded after its backward pass.

The integral term is evaluated separately in an unbatched manner. In the algorithm below, the gradients are all gradients of $\mcal{L}$; calculating gradients of the $\mathrm{NLL}$ training objective is a straightforward application of the chain rule, considering the $-1/N$ factor, penalty, and softplus transforms.

\begin{algorithm}
  \caption{Batched training of multivariate Hawkes processes (extension of Algorithm~\ref{alg:training_sequence})}
  \label{alg:batched_training_sequence}

  \SetKwBlock{Forward}{\textnormal{\textsc{Forward Pass}}}{}
  \SetKwBlock{Backward}{\textnormal{\textsc{Backward Pass}}}{}

  \textbf{define} $\mathrm{PrefixScan}_{\amul}(\,\cdot\,)$ \textbf{from} Section~\ref{sub:parallel_prefix_scan}\;

  \For{\textnormal{each epoch} $\ell$}{
    $\mcal{L}, \nabla\mu_p, \nabla\alpha_{p,q,k}, \nabla\gamma_k \gets 0$\;
    $\wt{\mb{R}}_0^k,\wt{\mb{L}}_0^k,\wt{\mb{K}}_0^k \gets (\mb{0}^{\top},1)^{\top}$\;

    \medskip

    \For{\textnormal{each batch} $b=1,\dots,B$}{
      \Forward{
        $\{\mb{P}_i^k\}_{i\in\mcal{I}_b} \gets \mathrm{PrefixScan}_{\amul}\left(\left\{(e^{-\gamma_k \Delta t_i}, e^{-\gamma_k \Delta t_i}\bm{\alpha}_{:,m_{i-1},k})\right\}_{i\in\mcal{I}_b}\right)$\;

        $\wt{\mb{R}}_i^k \gets \mb{P}_i^k \wt{\mb{R}}_{s_b}^k$, $\forall i\in\mcal{I}_b$\;

        $\{\mb{W}_i^k\}_{i\in\mcal{I}_b} \gets \mathrm{PrefixScan}_{\amul}\left(\left\{(e^{-\gamma_k \Delta t_i}, \Delta t_i\mb{R}_i^k)\right\}_{i\in\mcal{I}_b}\right)$\;

        $\wt{\mb{L}}_i^k \gets \mb{W}_i^k \wt{\mb{L}}_{s_b}^k$, $\forall i\in\mcal{I}_b$\;

        \medskip

        $\lambda_{m_i}(t_i) \gets \mu_{m_i} + \sum_{k=1}^{K} \gamma_k R_{m_i}^k(t_i)$, $\forall i\in\mcal{I}_b$\;

        $\mcal{L} \gets \mcal{L} + \sum_{i\in\mcal{I}_b} \log \lambda_{m_i}(t_i)$\;

        \medskip

        $D_i^k \gets R_{m_i}^k(t_i)-\gamma_k L_{m_i}^k(t_i)$, $\forall i\in\mcal{I}_b$\;

        \medskip

        \textbf{store} $\{\lambda_{m_i}(t_i)\}_{i\in\mcal{I}_b}$, $\{D_i^k\}_{i\in\mcal{I}_b}$ for backward pass\;
      }

      \medskip

      \Backward{
        $\{\mb{V}_i^k\}_{i\in\mcal{I}_b} \gets \mathrm{PrefixScan}_{\amul}\left(\left\{(e^{-\gamma_k \Delta t_i}, e^{-\gamma_k \Delta t_i}\mb{e}_{m_{i-1}})\right\}_{i\in\mcal{I}_b}\right)$\;

        $\wt{\mb{K}}_i^k \gets \mb{V}_i^k \wt{\mb{K}}_{s_b}^k $, $\forall i\in\mcal{I}_b$\;

        \medskip

        $\nabla\mu_p \gets \nabla\mu_p + \dfrac{\partial\mcal{L}}{\partial\mu_p}\Big|_{i\in\mathcal{I}_b}$ using Equation~\eqref{eq:grad_mu} and $\{\lambda_{m_i}(t_i)\}_{i\in\mcal{I}_b}$\;

        $\nabla\alpha_{p,q,k} \gets \nabla\alpha_{p,q,k} + \dfrac{\partial \mcal{L}}{\partial \alpha_{p,q,k}}\Big|_{i\in\mathcal{I}_b}$ using Equation~\eqref{eq:grad_alpha} and $\{\mb{K}_i^k\}_{i\in\mcal{I}_b}$\;

        $\nabla\gamma_{k} \gets \nabla\gamma_{k} + \dfrac{\partial \mcal{L}}{\partial \gamma_{k}}\Big|_{i\in\mathcal{I}_b}$ using Equation~\eqref{eq:grad_gamma} and $\{D_i^k\}_{i\in\mcal{I}_b}$\;

      }

      \medskip

      \textbf{retain} final states $\wt{\mb{R}}_{s_b+N_b}^k$, $\wt{\mb{L}}_{s_b+N_b}^k$, $\wt{\mb{K}}_{s_b+N_b}^k$ for the next batch\;
    }

    \medskip

    $\mcal{L} \gets \mcal{L}-\sum_{m=1}^{M}\int_0^T\lambda_m(t)\,dt$\;

    $\mu_p, \alpha_{p,q,k}, \gamma_k \gets \mathrm{OptimizerStep}(\mu_p, \alpha_{p,q,k}, \gamma_k, \nabla\mu_p, \nabla\alpha_{p,q,k}, \nabla\gamma_k, \mathrm{lr})$\;
  }
\end{algorithm}

\newpage
\section{Expanded Comparison of Intensity Implementations}
\label{app:comparison}

In this section, we extend the comparison of intensity implementations from Section~\ref{sec:results_comparison}. We test on a larger range of sequence lengths, from $N=2^3$ to $N=2^{26}$ (i.e., $N=8$ to $N\sim 6.7\times 10^7$), so that we can observe exact crossover points when certain implementations become most efficient. For all experiments, we use 64 cores of a Xeon 8580 processor, identical to the original experiment.

\begin{figure}[h]
  \centering
  \includegraphics[width=1\linewidth]{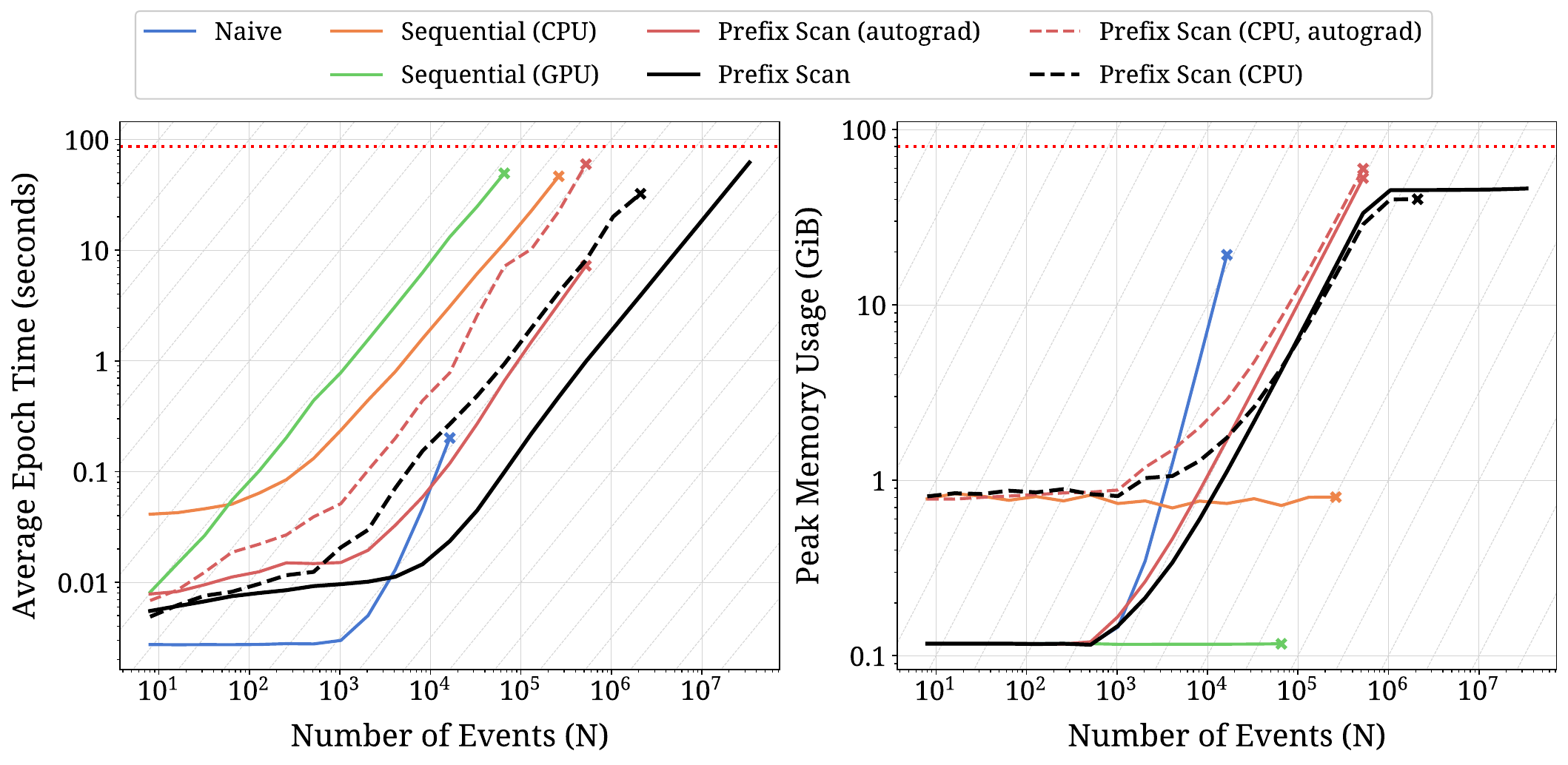}
  \caption{Our method on GPU (black, solid) and CPU (black, dashed), compared to other implementations of the intensity computation. Red dotted lines indicate the 24-hour walltime limit, as well as maximum available memory (80 GiB) for GPU runs.}
  \label{fig:appendix_performance_comparison}
\end{figure}

We first also consider our prefix scan approach when run on the CPU instead of the GPU (with either autograd or manual gradients). Figure~\ref{fig:appendix_performance_comparison} shows the results. As expected, the CPU implementation is slower than the GPU implementation due to lower parallelism, but it still substantially outperforms the sequential recurrence and other approaches.

\begin{figure}[h]
  \centering
  \includegraphics[width=1\linewidth]{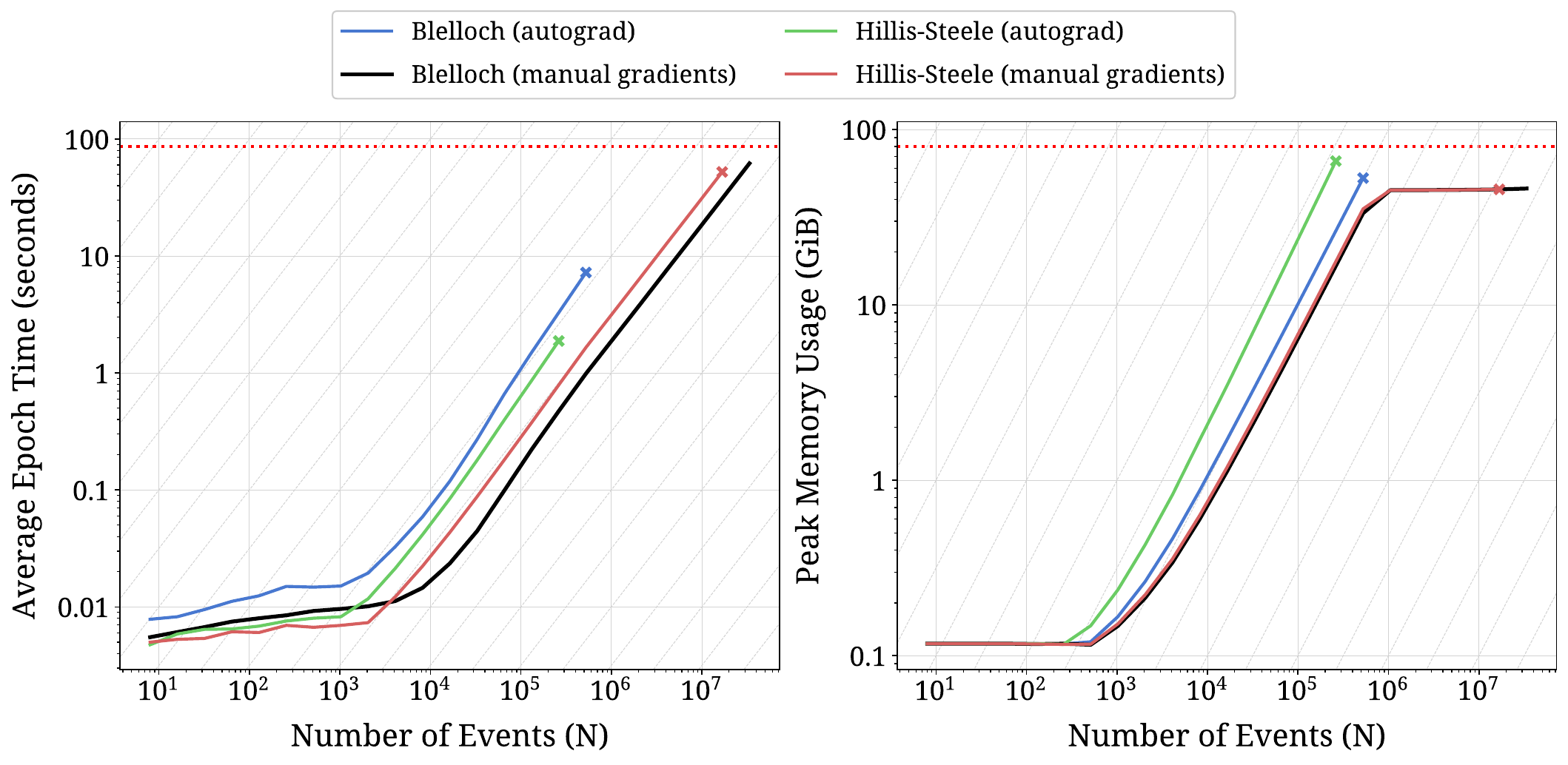}
  \caption{Performance of our method considering the Blelloch and Hillis-Steele scan implementations.}
  \label{fig:appendix_scan_comparison}
\end{figure}

Next, in Figure~\ref{fig:appendix_scan_comparison}, we compare whether the Blelloch or Hillis-Steele scan algorithm is better for implementing our method. Blelloch is faster when using manual gradients, which is to be expected. However, Hillis-Steele outperforms it under autograd gradients, as it generates a much simpler computation graph that is easier for autograd to traverse.

\newpage
\section{Correctness of Intensity, Log-Likelihood, and Gradients}
\label{app:correctness}

In order to demonstrate that our approach is exact in practice (up to floating-point numerical differences), we compare the computed intensity, log-likelihood, and parameter gradients against sequential and naive implementations. For each of $5$ random seeds, we simulate sequences of lengths $N\in\{10^2, 10^3, 10^4, 10^5, 10^6\}$ with $M=10$ nodes and $K=3$ decay kernels, resulting in $25$ total event sequences.

To each sequence, we fit a multivariate Hawkes process using the prefix scan calculations. At each epoch of the fit, we also compute the intensity and gradients using the naive and sequential implementations. Finally, we compute the index-wise relative errors between all three methods, and then take the maximum relative error across all indices, epochs, and seeds. For a computed value $x$ and its reference value $x^*$ (either the intensity, log-likelihood, or gradient), we calculate:
\begin{align*}
  \max_j \frac{|x_j - x^*_j|}{|x^*_{j}| + 10^{-12}}
\end{align*}
where $j$ indexes events or parameter indices.

Table~\ref{tab:correctness} reports the results, comparing our approach against the sequential recurrence (\ref{tab:correctness_a}), our approach against the naive implementation (\ref{tab:correctness_b}), and the sequential recurrence against the naive implementation (\ref{tab:correctness_c}). Note that the naive implementation could not be run past $N=10^4$ due to memory constraints, and so the corresponding entries are omitted. Across all parameter indices, epochs, sequence lengths, and seeds, comparing our implementation to the sequential recurrence, the maximum relative error in the intensities is $4.39\times 10^{-7}$ and in the parameter gradients is $6.68\times 10^{-5}$.

\begin{table}[h]
  \centering

  \caption{Comparison of different implementations for computing intensity, log-likelihood, and gradients. The maximum relative error is reported across all parameter indices, epochs, and seeds.}
  \label{tab:correctness}
  \medskip

  \begin{subtable}[t]{1\textwidth}
    \centering
    \caption{\normalsize Prefix scan vs. sequential recurrence}
    \label{tab:correctness_a}

    \begin{tabular}{cccccc}
      \toprule
      \textbf{$N$} & \textbf{Intensity ($\lambda_{m_i}(t_i)$)} & \textbf{Log-Lik.} & \textbf{$\mu$ gradient} & \textbf{$\alpha$ gradient} & \textbf{$\gamma$ gradient} \\
      \midrule
      $10^{2}$ & $2.34\times 10^{-7}$ & $3.58\times 10^{-7}$ & $2.45\times 10^{-7}$ & $5.10\times 10^{-6}$ & $2.74\times 10^{-6}$ \\
      $10^{3}$ & $2.45\times 10^{-7}$ & $6.15\times 10^{-6}$ & $3.28\times 10^{-7}$ & $8.95\times 10^{-7}$ & $5.77\times 10^{-6}$ \\
      $10^{4}$ & $3.21\times 10^{-7}$ & $7.92\times 10^{-5}$ & $1.01\times 10^{-6}$ & $2.59\times 10^{-6}$ & $4.09\times 10^{-6}$ \\
      $10^{5}$ & $3.86\times 10^{-7}$ & $6.48\times 10^{-4}$ & $3.38\times 10^{-6}$ & $8.75\times 10^{-6}$ & $1.14\times 10^{-5}$ \\
      $10^{6}$ & $4.39\times 10^{-7}$ & $4.50\times 10^{-3}$ & $9.91\times 10^{-6}$ & $6.68\times 10^{-5}$ & $3.83\times 10^{-5}$ \\
      \bottomrule
    \end{tabular}
  \end{subtable}

  \begin{subtable}[t]{1\textwidth}
    \centering
    \caption{\normalsize Prefix scan vs. naive implementation}
    \label{tab:correctness_b}
    \begin{tabular}{cccccc}
      \toprule
      \textbf{$N$} & \textbf{Intensity ($\lambda_{m_i}(t_i)$)} & \textbf{Log-Lik.} & \textbf{$\mu$ gradient} & \textbf{$\alpha$ gradient} & \textbf{$\gamma$ gradient} \\
      \midrule
      $10^{2}$ & $2.45\times 10^{-7}$ & $2.67\times 10^{-7}$ & $2.05\times 10^{-7}$ & $6.07\times 10^{-6}$ & $1.58\times 10^{-5}$ \\
      $10^{3}$ & $2.61\times 10^{-7}$ & $2.65\times 10^{-7}$ & $3.30\times 10^{-7}$ & $7.38\times 10^{-7}$ & $1.21\times 10^{-4}$ \\
      $10^{4}$ & $4.00\times 10^{-7}$ & $3.52\times 10^{-5}$ & $1.92\times 10^{-4}$ & $3.09\times 10^{-3}$ & $1.41\times 10^{-2}$ \\
      $10^{5}$ & --- & --- & --- & --- & --- \\
      $10^{6}$ & --- & --- & --- & --- & --- \\
      \bottomrule
    \end{tabular}
  \end{subtable}

  \begin{subtable}[t]{1\textwidth}
    \centering
    \caption{\normalsize Sequential recurrence vs. naive implementation}
    \label{tab:correctness_c}
    \begin{tabular}{cccccc}
      \toprule
      \textbf{$N$} & \textbf{Intensity ($\lambda_{m_i}(t_i)$)} & \textbf{Log-Lik.} & \textbf{$\mu$ gradient} & \textbf{$\alpha$ gradient} & \textbf{$\gamma$ gradient} \\
      \midrule
      $10^{2}$ & $2.45\times 10^{-7}$ & $3.40\times 10^{-7}$ & $2.92\times 10^{-7}$ & $6.19\times 10^{-6}$ & $1.63\times 10^{-5}$ \\
      $10^{3}$ & $3.11\times 10^{-7}$ & $6.15\times 10^{-6}$ & $2.03\times 10^{-7}$ & $1.10\times 10^{-6}$ & $1.25\times 10^{-4}$ \\
      $10^{4}$ & $3.90\times 10^{-7}$ & $8.96\times 10^{-5}$ & $1.92\times 10^{-4}$ & $3.09\times 10^{-3}$ & $1.41\times 10^{-2}$ \\
      $10^{5}$ & --- & --- & --- & --- & --- \\
      $10^{6}$ & --- & --- & --- & --- & --- \\
      \bottomrule
    \end{tabular}
  \end{subtable}

\end{table}


\end{document}